\documentclass[11pt,a4paper]{article}
\usepackage[hyperref]{acl2021}

\usepackage{times}
\usepackage{latexsym}

\usepackage{graphicx}
\usepackage{multirow}
\usepackage[normalem]{ulem}
\useunder{\uline}{\ul}{}
\usepackage{amssymb}
\usepackage{amsmath}
\usepackage{booktabs}
\usepackage{enumitem}
\usepackage{mathtools}
\usepackage{subcaption}
\usepackage{fontawesome} 
\usepackage{contour}
\usepackage{microtype}

\newcommand{\para}[1]{\smallskip\noindent\textbf{#1}}
\aclfinalcopy 

\title{Commonsense-Focused Dialogues for Response Generation: An Empirical Study}
\author{
Pei Zhou$^{1}$\thanks{~ Work done while Pei Zhou was an intern at Amazon Alexa AI} \quad Karthik Gopalakrishnan$^{2}$ \quad Behnam Hedayatnia$^{2}$ \quad Seokhwan Kim$^{2}$ \\ \textbf{Jay Pujara$^{1}$ \quad Xiang Ren$^{1}$ \quad  Yang Liu$^{2}$ \quad Dilek Hakkani-Tur$^{2}$}\\
$^1$ Department of Computer Science, University of Southern California\\
$^2$ Amazon Alexa AI\\
\small{\texttt{\{peiz,jpujara,xiangren\}@usc.edu},} \\
\small{\texttt{\{karthgop,behnam,seokhwk,yangliud,hakkanit\}@amazon.com}}
}

\date{}

\begin{document}
\maketitle

\begin{abstract}
Smooth and effective communication requires the ability to perform latent or explicit commonsense inference.
Prior commonsense reasoning benchmarks (such as SocialIQA and CommonsenseQA) mainly focus on the discriminative task of choosing the right answer from a set of candidates, and do not involve interactive language generation as in dialogue. Moreover, existing dialogue datasets do not explicitly focus on exhibiting commonsense as a facet. In this paper, we present an empirical study of commonsense in dialogue response generation. We first auto-extract commonsensical dialogues from existing dialogue datasets by leveraging ConceptNet, a commonsense knowledge graph. Furthermore, building on social contexts/situations in SocialIQA, we collect a new dialogue dataset with 25K dialogues aimed at exhibiting social commonsense in an interactive setting.
We evaluate response generation models trained using these datasets and find that models trained on both extracted and our collected data produce responses that consistently exhibit more commonsense than baselines. Finally we propose an approach for automatic evaluation of commonsense that relies on features derived from ConceptNet and pretrained language and dialog models, and show reasonable correlation with human evaluation of responses' commonsense quality.
We are releasing a subset of our collected data, {\bf Commonsense-Dialogues}\footnote{The released version was manually validated by in-house crowd workers and hence is slightly different from the initial version used in the paper. URL:
\href{https://github.com/alexa/commonsense-dialogues}{github.com/alexa/commonsense-dialogues}}, containing about 11K dialogs.

\end{abstract}

\section{Introduction}\label{intro}

Open-domain dialogue response generation (RG) models aim to provide human-like natural language responses given dialogue histories~\cite{chen2017survey}. 
To improve generated response quality, many studies have been conducted to develop knowledge-grounded RG~\cite{ghazvininejad2018knowledge, gopalakrishnan2019topical}, personalized dialogue agents~\cite{zhang2018personalizing}, empathetic response~\cite{rashkin2019towards}, etc.
For all the above-mentioned directions for RG, large-scale dialogue data geared towards the specific goals is crucial, since most current state-of-the-art neural RG models require training on appropriate and large data. 
Therefore several datasets have been collected to support such research efforts such as knowledge-grounded dialogues~\cite{ghazvininejad2018knowledge, gopalakrishnan2019topical}, PersonaChat~\cite{zhang2018personalizing}, and EmpatheticDialogues~\cite{rashkin2019towards}. 
Producing natural and logically-coherent responses given dialogue contexts involves making commonsense inferences during the communication. For example, if someone says ``\emph{I'm going to perform in front of a thousand people tomorrow...}'' the listener is likely to conclude that the speaker is probably feeling nervous and respond by comforting them: ``\emph{Relax, you'll do great!}''
In contrast to other efforts to make RG models more empathetic or knowledgeable, there is a lack of commonsense focused dialogue data for both training neural models and evaluation. An ideal dataset for studying commonsense in RG needs to simulate how humans have multi-turn conversations as much as possible. Existing commonsense-focused work in RG uses extracted post-response pairs from Reddit~\cite{zhou2018commonsense}, which are single-turn and rough approximations for real-life conversations.

Aiming to bridge the gap in commonsense for dialogue response generation, we collect a large-scale multi-turn open-domain dialogue dataset that is focused on commonsense knowledge. 
We first consider extracting \emph{commonsense-focused} dialogues from three existing dialogue datasets by identifying responses that contain commonsense inferences using ConceptNet~\cite{liu2004conceptnet}. This filtering results in 21k dialogues. Then we collect 25k new dialogues focusing on social commonsense inferences, where prompts are context sentences describing an event in the SocialIQA data~\cite{sap2019social}. 

To study commonsense in RG, we train large generative language models on our datasets and compare with models trained on existing datasets. We find through sampled human evaluation that our dataset helps to generate more commonsensical responses (average score of 6.9 out of 10 compared to 4.8 using other data), and automatically generated responses still have a large gap in comparison to human performances (9.2 out of 10).
To help lower the evaluation cost and increase the efficiency of evaluating commonsense in RG, we further propose an automatic metric using combined neural and symbolic features derived from ConceptNet, and show that this metric has reasonable correlation with human annotations and symbolic features contribute positively to system performance.

Our contributions are as follows: (1) We create the first large-scale open-domain dialogue dataset focusing on social commonsense inferences. This includes a new collection of 25k dialogues based on SocialIQA event prompts, and ConceptNet filtered data from some existing data sets.  We are also releasing the ConceptNet filtered portion of our data collection, Commonsense-Dialogues, which contains about 11K dialogs. (2) We benchmark our dataset and show that models trained on our dataset helps make models produce more commonsensical responses. (3) We propose the first automatic metric for evaluating the commonsense plausibility in response generation that reaches statistically significant correlation with human annotations.

\section{Task Introduction and Motivations}\label{task}
\subsection{Commonsense-Focused Dialogue Response Generation}
We study commonsense-focused response generation for dialogues. Commonsense can be defined as ``the basic level of practical knowledge and reasoning concerning everyday situations and events that are commonly shared among most people''~\cite{sap2020commonsense}. Dialogue response generation is the task of generating a response turn $r$ in a conversational setting given previous history turns $h$. Thus by combining these two together, we want to examine models' ability to produce responses that make sense or is plausible in terms of commonsense.

\subsection{Motivations}
\paragraph{Lack of Commonsense-Focused Analysis on Existing Dialogue Datasets}
Numerous dialogue data has been collected for training RG models and other dialogue-related tasks. As mentioned before, many different aspects of RG have been explored, such as knowledge-grounded~\cite{ghazvininejad2018knowledge, gopalakrishnan2019topical} and empathy~\cite{rashkin2019towards}, whereas, to the best of our knowledge, there is no study or large-scale multi-turn data for analyzing whether model-generated responses present the ability to communicate with commonsense knowledge or reasoning.

\paragraph{Lack of real-life interactive setting for Commonsense Reasoning Benchmarks}
Current commonsense reasoning (CSR) benchmarks mostly target models' ability to choose a right answer from several candidates given a question. We argue that this is a highly artificial scenario as models do not get options to choose from in real-life, and often they need to generate utterances. Recent work such as CommonGen~\cite{lin2020commongen} has started to explore generative settings to examine commonsense in natural language processing (NLP) models. This line of work, however, is still far from real use cases as it does not consider a real-life interaction task setup such as conversations. Thus we argue that existing commonsense benchmarks in NLP are not enough to train a language agent that produces smooth interpersonal communications, nor evaluate whether models have such capabilities.

\section{Commonsense Focused Dialogue Collection}\label{prob}
To collect more commonsense focused dialogues for response generation model training and evaluation, our effort is along two directions: filtering existing data to collect dialogues with responses that consist of commonsense (Section~\ref{sec3_filter}), and curating new data using prompts from a commonsense reasoning multiple-choice benchmark SocialIQA (Section~\ref{sec3_newdata}).

\subsection{Filtering Based on Existing Dialogue Datasets}\label{sec3_filter}
We propose a simple process for filtering commonsense in dialogues and present our analysis of three dialogue datasets with different focuses: DailyDialog~\cite{li2017dailydialog}, EmpatheticDialogues~\cite{rashkin2019towards}, and MuTual~\cite{cui2020mutual}.
The general idea is to refer to a commonsense knowledge graph (CSKG) such as ConceptNet~\cite{liu2004conceptnet} to identify potential commonsense triples ($e_1, r, e_2$) expressing a commonsense assertion between turns in a dialogue. 
The following describes the detailed process. 

\para{Identify Candidate Concepts}
The first step is to identify potential candidates for concept entities in the commonsense triples. For a turn in a dialogue, we use a part-of-speech (POS) tagger to find the nouns, verbs, and adjectives that are not stopwords and then construct a set of potential concepts by including the lemmatized version of these words. We use the POS tagger, lemmatizer, and stopword list from the Natural Language Toolkit (NLTK) package~\cite{bird2009natural}. This step results in a set of concept words for \emph{each turn} of a dialogue. For example, consider an exchange between two participants in a conversation: ``Hi, I want to find a doctor”, “What kind of doctor are you looking for? A general doctor or a specialist?”, the concept sets for the two turns are ``\emph{want, find, doctor}'' and ``\emph{look, general, doctor, specialist}'', respectively.

\para{Query ConceptNet for Neighboring Entities}
With a set of concepts we extract for every dialogue turn, we then identify a list of candidate triples $(e_1, r, e_2)$ expressing commonsense assertions about each concept such that we can later check if some of those assertions indeed appear in this dialogue. We rely on the widely-used ConceptNet~\cite{liu2004conceptnet} as the knowledge resource, which consists of commonsense knowledge about various concepts. Specifically we use the ConceptNet containing single-word concepts pre-processed by~\citet{zhou2018commonsense}. For each concept we identified in a turn, we store all triples in ConceptNet that contain this concept, either as subject or object. Using the above example, example triples about ``doctor'' include ``doctor LocateAt hospital'', ``patient RelatedTo doctor'', and ``specialist TypeOf doctor''.

\para{Search Entities in the Next Turn}
After getting a list of commonsense triples $(e_1, r, e_2)$ containing concepts in a particular turn using ConceptNet, we next examine if any of the \emph{other} entity in the triples appears in the concept set of the next turn. 
In the example dialogue exchange above, where ``doctor'' is a concept appearing in a turn,
for the triple ``specialist TypeOf doctor'',  we search if ``specialist'' is in the concept set of the next turn. 
Since we find such a match, 
we record this triple to be a commonsense assertion that might be implied in the response. 


\para{Filtering Results}\label{filter_results}
We filter dialogues using the above-mentioned approach: if we can successfully find a matching triple between two adjacent turns, we keep the dialogue as it might contain commonsense assertions identified from ConceptNet. We consider three dialogue datasets in this study:
\begin{itemize}
    \item DailyDialog(DD)~\cite{li2017dailydialog}. It includes general-domain day-to-day dialogues crawled from various English learning websites.
    \item EmpatheticDialogues (ED)~\cite{rashkin2019towards}. It is an empathy-focused dialogue dataset crowdsourced from Amazon Mechanical Turk (MTurk).
    \item MuTual~\cite{cui2020mutual}.   It is a reasoning-focused response selection dataset based on English listening comprehension exams for Chinese students.
\end{itemize} 
We choose these three datasets to examine three different types of focuses in dialogue datasets: general-domain, empathy, and general reasoning (but not specifically on commonsense).

After the process, we find that in the training sets, around 7k out of the 11k dialogues (63\%) from Dailydialogue contain at least one matched triple between their turns, and 9.5k out of the 18k for   EmpatheticDialogues (53\%),  and 5k out of 7k (73\%) for MuTual dialogues.
For the valid and test sets, the proportion of such dialogues is similar to that in the training sets for these three data sets.  

Note that there are some limitations in our ConceptNet based data selection approach. 
First, we match concept entities based on just surface form, rather than semantic meaning or word senses in the context.  
Second, we are only using single word concepts, not phrases. 
Third, we are only considering one-hop concept relation identified in ConceptNet.  
The first one may affect the precision of the selected dialogues, and the other two reasons affect the recall.  
Without human annotated commonsense reasoning for dialog turns, we can not compute the exact performance of our filtering method. 
We plan to conduct some human annotation in our future work. 
Among the three data sets used in this study, the fact that there is a higher percentage of dialogues selected in MuTual may indicate that data focuses more on reasoning and thus is more likely to contain commonsense relations. 

\subsection{New Data Collection Using SocialIQA Prompts}
\label{sec3_newdata}
To facilitate commonsense-guided response generation training, we collect more dialogues with a focus on getting responses that require commonsense. Specifically, we make use of an existing commonsense multiple-choice benchmark SocialIQA~\cite{sap2019social} to crowdsource dialogues. This section provides background on SocialIQA, the crowdsourcing process, and the resulting dialogues.

\para{Background and motivation}
We collect dialogues by prompting crowdsourcing workers on Amazon Mechanical Turk (MTurk) with context sentences from SocialIQA that describe an event in everyday social scenarios. 
SocialIQA~\cite{sap2019social} is a large-scale commonsense reasoning benchmark about social situations. It contains around 38k multiple-choice questions, each consisting of a context sentence, a question, and three answer choices. Context was generated by rewriting events from ATOMIC~\cite{sap2019atomic}, a large knowledge graph (KG) that contains inferential knowledge about the causes and effects of 24k short events. An example event in ATOMIC is ``PersonX spills all over the floor'', which crowd workers were asked to turn into a sentence by adding names, fixing potential grammar errors, and filling in placeholders, resulting in a context like ``Alex spilled food all over the floor.''

We choose to use SocialIQA contexts because of three reasons: (1) they are specific instantiations of the event phrases found in the knowledge graph ATOMIC, which guarantees that there is at least one potential commonsense inference that can be made from the event; (2) ATOMIC covers a wide range of commonsense motivations and reactions and thus the contexts also embed diverse commonsense; (3) the rewriting process from SocialIQA ensures that the context sentences are well-formed and similar to natural sentences, which we expect is not hard for crowd workers to come up with a dialogue.

\para{Prompt selection}
We inspected around 200 contexts trying to write a dialogue and found that the contexts that we had the most difficulty with are the ones that are too short or do not contain an interesting event to start a conversation. For example, contexts such as ``Robin stopped eating the food to save room for dessert'' might not be an interesting event to talk about in a dialogue. 
To select appropriate contexts as prompts for dialog writing, we apply a simple heuristic criteria: the context has to be either longer than 15 words or contains a punctuation such as a comma or a period in the middle. The intuition is that longer contexts are easier to write a dialogue with because they contain more information and a punctuation often indicates a development in the narrative of the event (e.g., ``Tracy performed her function. Their employer gave them a raise''). This makes the event more complicated, and thus avoids too trivial events. We also filter out context sentences that do not contain any person names.
As a result of this preprocessing, we kept 12.8k out of 33k contexts in the training set and 754 out of 2k contexts in the development set, adding up to 13.5k contexts from SocialIQA.

\begin{table*}[]
\resizebox{\textwidth}{!}{
\begin{tabular}{|c|l|}
\hline
\textbf{Prompts}                                                                                                                                            & \multicolumn{1}{c|}{\textbf{Dialogue Examples}}                                                                                                                                                                                                                                                                                                                                                                                                                                                                                                                                                                                               \\ \hline
\multirow{2}{*}{Tracy performed her function.}                                                                                                              & \begin{tabular}[c]{@{}l@{}}Tracy: I got a raise today. Totally unexpected. \\            My boss told me I was doing a great job.\\ Friend: It feels good to be rewarded for hard work.\\ Tracy: I've been trying my best at this job.  I've been \\            putting in long hours to make sure I get everything done.\\ Friend: Sounds like your boss recognized that.\\ Tracy: It's great when people can work well together.\end{tabular}                                                                                                                                                                                               \\ \cline{2-2} 
                                                                                                                                                            & \begin{tabular}[c]{@{}l@{}}Tracy: Get dressed. We're going out to celebrate my raise.\\ Friend: Awesome. What did your boss say when you got it?\\ Tracy: She said I did my job very well and deserved it.\\ Friend: You should be so proud. You've earned it.\end{tabular}                                                                                                                                                                                                                                                                                                                                                                   \\ \hline
\multirow{2}{*}{\begin{tabular}[c]{@{}c@{}}Addison wanted to go on a trip to Mexico, \\ and messaged all of his friends to set up a schedule.\end{tabular}} & \begin{tabular}[c]{@{}l@{}}Addison: Hey guys! I'm planning a Mexico vacation for everyone! \\                Let's work out a schedule so we can all do somethings we \\                want to do together.\\ Friend: I'm down! We should get in some scuba diving. I've been \\             wanted to get some good underwater photos for my gallery.\\ Addison: That sounds fun! I've never scuba dived before. Do you \\                 have to have any training?\\ Friend: They give you a little course on how to use the equipment. \\             You can opt out and just do the snorkeling if it's too intimidating.\end{tabular} \\ \cline{2-2} 
                                                                                                                                                            & \begin{tabular}[c]{@{}l@{}}Addison: I think we'll go to Mexico next.\\ Friend: That sounds exciting. Did you find a time that works for everyone.\\ Addison: No! But I'm going to message them right now to find out!\\ Friend: Yeah, You had better figure out a time as soon as possible.\\              Scheduling is super hard with more than 3 people.\\ Addison: Yep. But we'll get it done! My friends are the best at this!\end{tabular}                                                                                                                                                                                             \\ \hline
\end{tabular}
}
\caption{Examples for prompts from SocialIQA and generated dialogues from crowdsourcing on MTurk.}
\label{Tab:dial_examples}
\end{table*}

\para{Dialogue Collection}
Using selected contexts from SocialIQA, we ran a task on MTurk asking each worker to write a dialogue with 4 to 6 turns between two friends about the event described in the context. Note that, this is a `self-talk' dialog collection. Specifically, since there is a name appearing in the context, we ask a worker to write a dialogue by first imagining that they are the person mentioned in the context and are talking with their friend about the event described. For example, consider the context above (``Tracy performed her function. Their employer gave them a raise''), we ask a worker to imagine themselves to be ``Tracy'' and that they are talking to a friend (also played by themselves) about getting a raise. 

We pose three requirements for turkers in order to work on our task: locate in US, UK, or Canada; successful HITS are over 1000, and with more than 95\% HIT acceptance rate. We pay MTurk workers \$0.5 for each instance, roughly translating to 10 dollars per hour, well above the minimum wage of US.  
 
To account for multiple plausible dialogues expanded from the context event, we assign each context to five different MTurk workers.  
We randomly sample 5k context sentences out of 13.5k filtered ones and collect five dialogues for each context, resulting in 25k dialogues.
Examples of our collected dialogues are shown in Table~\ref{Tab:dial_examples}.

For our collected data, we follow the same filtering steps as used for other existing data (Section ~\ref{sec3_filter}).
This ConceptNet filtering identifies about 11K dialog from the entire collection.\footnote{Our released data is based on these ConceptNet filtered conversations. We recruited in-house crowd workers to manually check the dialogs for profanity, speaker name, and other issues in the data. Note the experiments conducted in this paper used the initial collection, not this released version.} 
Though we expect the SocialIQA contexts are from ATOMIC and may trigger more commonsensical dialogue, we find this is not the case since the percentage of dialogues containing ConceptNet triples is even lower than what we observed for the other existing data sets. 
This may be because of the limitations of the filtering method we are using as described earlier: matching to ConceptNet is based on surface textual form and concepts are on word-level, which omits deeper and more contextual commonsense relationships.






\section{Experiment Setup and Evaluation Methods}\label{exp}

The focus of this study is to examine how commonsense plays a role in dialogue response generation. In previous sections, we propose a simple filtering method to obtain \emph{commonsense-focused} dialogues from existing three datasets and crowdsource more dialogues based on the SocialIQA commonsense reasoning benchmark. Here we aim to evaluate response generation models' ability to produce responses that follow commonsense and if training on \emph{commonsense-focused} dialogue data helps boost model performance. 
In addition to using automatic referenced metrics and human evaluation, we also propose a new automatic unreferenced metric aiming to evaluate responses for commonsense quality.  

\subsection{Experiment Settings}
For response generation models, we take one of the state-of-the-art pre-trained language models, GPT2~\cite{radford2019language}, and further train it on our training data sets. Specifically, the model is trained in a multitask fashion that minimizes the LM loss as well as the multiple choice loss following~\citet{transfertransfo}, and generates responses for a given dialog history.

We consider the follow three types of training data setups. 
\begin{itemize}
    \item Existing data sets, including DailyDialog~\cite{li2017dailydialog} (DD),  EmpatheticDialogues~\cite{rashkin2019towards}(ED), and
    Topical-Chat~\cite{gopalakrishnan2019topical}, a knowledge-grounded open-domain dataset with around 11k dialogues. MuTual~\cite{cui2020mutual} is not included since it is designed for response selection. 
    
    \item 
    As described in Section~\ref{filter_results}, we use ConceptNet to search for potential triples in response turns and filter three dialogue datasets, DD, ED, and MuTual. We combine the three filtered dialogues from these datasets to form our training data, named `filter existing' (FE, total around 21K dialogues).
    
    \item  The third category includes our collected dialogues using SocialIQA contexts. This is used along with the FE data above: FE and all of the 25k collected dialogues (FE+new crowdsourced), and FE plus the 11K filtered dialogues of our collected data (FE+filtered crowdsourced).


\end{itemize}

To evaluate models' response generation capabilities, we sample 10\% of the FE+new data, resulting in 4.6k testing dialogues with no overlap with the training set of any of the settings above. 
We use GPT2 trained on different versions of dialogue data (6 trained GPT2 models in total) to generate a randomly sampled response for each turn of our test set dialogues. 


\subsection{Evaluation Metrics} 
We perform automatic evaluation on the test set and human evaluation on sampled dialogs. 

\para{Automatic Evaluation}
We consider several widely-used automatic metrics for evaluating response generation: perplexity of the reference responses in the data, Meteor score~\cite{banerjee2005meteor}, ROUGE score~\cite{lin2004rouge}, and BERTScore~\cite{zhang2019bertscore}. Note that these metrics (except perplexity) provide general evaluation of the generated responses, but do not specifically focus on commonsense plausibility.

\para{Human Evaluation}
Since there is no existing evaluation method that reliably examines whether a response follows commonsense and correlates with human judgements, we ask humans to score system generated responses as well as the reference response given a dialogue history. We sample 300  history-response pairs from dialogues in our test set to perform human evaluation.  
All the model-generated responses from the 6 trained models above and the original response (human response) (around 2100 responses in total) are scored in terms of \emph{commonsense plausibility} by MTurkers. 
We specifically asked workers to score the responses in terms of \emph{commonsense plausibility} using a scale of 1 to 10. We also instructed them that criteria such as grammatical correctness and fluency should not be taken into much account and they should focus on evaluating the commonsense aspect of the response. 
Three annotators evaluated each response. 
We calculate the average human scores and variance to measure the performances of different responses. 


\subsection{Proposed Automatic Metric for Commonsense}
Human evaluation is expensive to obtain, especially when the dataset is large. In addition, they are also subjective and hard to reproduce. Aiming to provide a reliable and scalable automatic metric focusing on commonsense in response generation, we propose an unreferenced automatic metric, which is a regression model trained from the human annotation scores for different responses. The metric is reference-free, meaning that it does not require human ground truth response when scoring a model-generated response, unlike referenced metrics such as BLEU, ROUGE, Meteor.  

\para{Regressor model}
We use a simple multi-layer perceptron (MLP) as our regressor and consider both neural and symbolic features to train the MLP model. 
For symbolic features, we consider the number of one-hop and two-hop triples that can be found between the dialogue history and the response turn from ConceptNet. The triple identifying process is the same as our filtering process described earlier (Section~\ref{sec3_filter}). That is, we first identify a set of concepts in the response turn and query ConceptNet for potential triples and match those with the other concepts appearing in the dialogue history. Two-hop triples are searched in a similar manner, with the only difference being that the number of potential triples will be much larger. We also include the length of the response as an additional feature.
As for neural features, we use the scores from a dialogue-focused language model DialoGPT~\cite{zhang2020dialogpt} on both the response itself and the dialogue history concatenated with the response. The score from DialoGPT can be considered as the plausibility of the sentence. We train this MLP model using the human evaluation scores for different responses.


\section{Results and Analysis} \label{result}
\subsection{Automatic Evaluation Results}
Table~\ref{tab:automatic_results} shows results according to automatic metrics on our 4.6K testing dialogues. We find that perplexity scores for the GPT2 model trained on filtered existing dialogue data (FE), or plus new collected data (FE+Crowdsourced), are much lower than that just trained on existing datasets as is. 
There are several reasons for this. 
One is that since the testing dialogues are from the filtered version, training on those better matches the evaluation scenario. 
In addition, the test set is a sample of multiple data sets, and thus training on just one data set does not perform well. 
Finally the combined data (the last three rows in the table) is larger in size (see training size in Table~\ref{tab:human_results}).  
However, note the gain from the increasing training data size decreases in comparison to the difference between using the filter data settings and those single data sets. 
Meteor and ROUGE scores for all the trained models are quite low, and show less differences, probably indicating the limitation of these metrics for dialog response evaluation. 
BERTScore shows a similar pattern as perplexity in terms of model quality.


\begin{table}[ht]
\resizebox{\columnwidth}{!}{
\begin{tabular}{c|c|c|c|c}
\toprule
Data              & Perplexity &  Meteor    & ROUGE & BERTScore        \\ \midrule
DD                & 31.25           & 0.06          & 0.06 & 0.12         \\
ED                & 24.80           & 0.08          & 0.08 & 0.14         \\
TC                & 28.48           & 0.09          & 0.08  & 0.11         \\
Filtered Existing (FE)        & 13.20           & 0.09      & 0.08    & 0.16         \\
FE+Crowdsourced      & 11.31           & 0.09 & 0.08 & 0.17 \\
FE+Filtered Crowdsourced & 12.27           & 0.09  & 0.08 & 0.17 \\  \bottomrule
\end{tabular}
}
\caption{Automatic evaluation results for different models on the test set.}
\label{tab:automatic_results}
\end{table}

\subsection{Human Evaluation Results}\label{sec:human eval}
Table~\ref{tab:human_results} shows the human evaluation scores on 300 responses for models trained with different types of data.
The most obvious and perhaps expected finding is that GPT2, no matter trained on what types of data, is still way behind human performance (6.86 with high variance versus 9.3 with low variance). 
By analyzing different variables that cause performance difference, we find the following patterns, some of which are similar to using automatic metrics. 
(1) Using the Filtered Existing dialogue data (FE) helps improve the average of commonsense scores (more than 1 point improvement comparing to using individual data sets), but variance remains high; (2) Including our collected dialogues further increases the average (FE+Crowdsourced), and also decreases the variance in response quality in terms of commonsense plausibility; (3) Regarding our collected data, using the filter subset of it yields slightly better performance than using the entire data collection. 
This suggests that even though our data is collected using SocialIQA events, some dialogues may not be commonsense rich, which is also reflected by the percentage of dialogues that contain ConceptNet triples as discussed earlier. 
In addition, it shows that though overall increasing training data size benefits model performance, the quality of data plays a more important role.  
We plan to perform more sophisticated data selection and commonsense annotation for our data set in the future.  
We include examples of responses from humans and models trained on these different types of data as well as annotation scores in Appendix~\ref{sec:appendix} Table 5. 
It shows some different characteristics of the responses, for example, empathy in the responses using ED model, and richer information (though inappropriate since they are off topic) using TC model.




\begin{table}[ht]
\resizebox{\columnwidth}{!}{
\begin{tabular}{c|c|c|c}
\toprule
Data              & Training Size & Avg. Score     & Variance       \\ \midrule
DD                & 11k           & 4.677          & 11.977         \\
ED                & 18k           & 4.998          & 12.233         \\
TC                & 10k           & 4.558          & 11.562         \\
Filtered Existing (FE)         & 21k           & 5.968          & 12.426         \\
FE+Crowdsourced      & 46k           & \textbf{6.767} & \textbf{9.067} \\
FE+Filtered Crowdsourced & 31k           & \textbf{6.865} & \textbf{8.684} \\ \midrule
Human response             & N/A           & \textbf{9.298} & \textbf{2.544} \\ \bottomrule
\end{tabular}
}
\caption{Average human scores and variance on human responses and system generated responses from GPT2 models trained on different data.}
\label{tab:human_results}
\end{table}

\subsection{Proposed Commonsense Automatic Evaluation Results}
We now examine the correlation of our proposed automatic metric (MLP regressor) with human scores on the testing portion of our annotations. We cross-validate on the collected dialogues with 0.8/0.1/0.1 proportions. For comparison, we consider three baselines: our MLP with only symbolic features, our MLP with only neural features, and FED~\cite{mehri2020unsupervised}, which uses DialoGPT to score how likely the \emph{next} turn after the response expresses confusion. It requires no training nor human references, and has been shown to correlate with humans judgements on different criteria (commonsense not included). 
Table~\ref{tab:metric_results} shows the Spearman's correlation of the system computed scores and human annotation scores using all the annotated data in a cross-validation setup.
We can see that our simple MLP-based regressor reaches the highest spearman's correlation with human scores, outperforming other baselines significantly.
However, such a correlation result still suggests a large gap for a reliable scorer targeting commonsense evaluation for dialogue response generation. We also notice that FED performs poorly in terms of commonsense evaluation. Furthermore, there is a large correlation drop when considering either symbolic or neural features alone in our model, indicating that they might each capture a different aspect for evaluating commonsense.

\begin{table}[ht]
\resizebox{\columnwidth}{!}{
\begin{tabular}{c|c|c|c}
\toprule
\multicolumn{2}{c|}{\textbf{Metrics}} & \textbf{Spearman's Correlation} & \textbf{p-Value}  \\ \midrule
\multicolumn{2}{c|}{FED}        & -0.00797                      & 0.80569           \\ \hline
 & Symbolic    & 0.12336                       & 1.27E-08          \\
Ours & Neural      & 0.06176                       & 0.00450           \\
& All features        & \textbf{0.20789}              & \textbf{4.53E-22} \\ \bottomrule
\end{tabular}
}
\caption{Spearman's correlation and p-values for different automatic metrics with human scores.}
\label{tab:metric_results}
\end{table}


\section{Related Work}\label{rel_work}
\subsection{Commonsense Reasoning}
The majority of recent commonsense reasoning benchmarks~\cite{zellers2018swagaf, talmor2019commonsenseqa,bisk2019piqa,sap2019social} test a model's ability to choose the correct option given a context and a question; pre-trained language models have reached high performance on these benchmarks after fine-tuning. 
There have been many benchmarks that focus on reasoning abilities in multiple tasks such as reading comprehension~\cite{huang2019cosmos, yu2020reclor}, dialogue systems~\cite{cui2020mutual}, and natural language inference~\cite{MNLI}, which involve inferences on language.
Recent work also aims to probe models in these tasks to see if reasoning is actually achieved~\cite{richardson2020does,richardson2020probing,zhou2020rica}.
In this study we tackle the response generation problem in dialogues, with a focus on collecting commonsense rich dialog data and evaluating commonsense quality of model responses.

\subsection{Open Domain Dialogue Response Generation}

Recently open domain dialog systems have been modeled using end-to-end approaches, more specifically encoder-decoder architectures~\citep{sordoni2015neural, serban2017hierarchical, serban2016building, vinyals2015neural}. Recent work focused on finetuning large pre-trained transformer models~\cite{radford2019language,zhang2020dialogpt} on dialog data. 
Many dialog datasets have been collected with different focuses such as incorporating knowledge~\citep{gopalakrishnan2019topical,dinan2018wizard}, empathy~\cite{rashkin2019towards}, task completion~\cite{budzianowski2018multiwoz}, consistency~\cite{nie2020like}, personality~\cite{zhang2018personalizing} and reasoning~\cite{cui2020mutual} within dialog systems. 
There has also been work on combining a variety of datasets to exhibit multiple attributes~\cite{roller2020recipes}.

\subsection{Dialog Response Evaluation}
Due to the diverse responses that a dialog system can output, referenced automatic metrics (such as BLEU, ROUGE, Perplexity) do not correlate well with human judgement of these systems~\citep{deriu2020survey, liu2016not}. As a result, human evaluation has become the de-facto standard to evaluate dialog systems. However human evaluation is costly.
Recently model-based metrics have been proposed with good correlation with human annotations~\citep{zhang2019bertscore, sellam2020bleurt, mehri2020usr, mehri2020unsupervised, tao2018ruber, lowe2017towards}. Most metrics focus on evaluating the coherence or appropriateness of a response with respect to its dialog context.~\cite{mehri2020unsupervised} identified 18 different dialog qualities such as interesting and topic depth. However none of these metrics evaluate the commonsense of a response, which is the focus of this work.

\section{Conclusion}\label{conclusion}
We present our empirical study on commonsense in dialogue response generation. To obtain data for commonsense-focused analysis in open domain response generation, we use two strategies: filtering existing dialogue data using a commonsense knowledge graph ConcepetNet, and collecting new dialogues using prompts from multiple-choice commonsense benchmark. Our data has a few limitations such as our filtering process focuses on word-level matching to ConceptNet, which might omit more complex commonsense relations embedded in dialogues. We leave deeper analysis of how implicit commonsense is represented in dialogues and how to elicit multi-hop granular reasoning steps during communications to future work.
Our experimental results show that our newly collected data helps boost response generation model performance based on human evaluation of commonsense. To close the gap in automatic evaluation metric for response generation, we propose a simple MLP regressor using both neural and symbolic features, and show promising correlation with human judgements. 

We are releasing the ConceptNet filtered portion of our data collection, with further manual examination, {\bf Commonsense-Dialogues}, which consists of about 11K dialogs. 
We hope our work and this new data will help with future attempts to make models produce responses with more commonsense, which is a challenging but crucial task to tackle in dialog systems.  


\section*{Ethics and Broader Impact}
Our work uses ConceptNet to filter for commonsense-focused dialogues, but ~\citet{mehrabi2021lawyers} have found representational harms in common sense resources. We acknowledge that the generated responses from models we use might contain biases.
All of the dialogue datasets and models are in English, which benefits English speakers more. We used Amazon Mechanical Turks for data collection. We pay turkers around \$14 per hour, well above the highest state minimum wage and engage in constructive discussions if they have concerns about the process. We also give each annotation instance enough time so that we do not pressure annotators.

\bibliography{anthology,acl2021_old}

\begin{thebibliography}{45}
\expandafter\ifx\csname natexlab\endcsname\relax\def\natexlab#1{#1}\fi

\bibitem[{Banerjee and Lavie(2005)}]{banerjee2005meteor}
Satanjeev Banerjee and Alon Lavie. 2005.
\newblock Meteor: An automatic metric for mt evaluation with improved
  correlation with human judgments.
\newblock In \emph{Proceedings of the acl workshop on intrinsic and extrinsic
  evaluation measures for machine translation and/or summarization}, pages
  65--72.

\bibitem[{Bird et~al.(2009)Bird, Klein, and Loper}]{bird2009natural}
Steven Bird, Ewan Klein, and Edward Loper. 2009.
\newblock \emph{Natural language processing with Python: analyzing text with
  the natural language toolkit}.
\newblock " O'Reilly Media, Inc.".

\bibitem[{Bisk et~al.(2020)Bisk, Zellers, Bras, Gao, and Choi}]{bisk2019piqa}
Yonatan Bisk, Rowan Zellers, Ronan~Le Bras, Jianfeng Gao, and Yejin Choi. 2020.
\newblock Piqa: Reasoning about physical commonsense in natural language.
\newblock \emph{AAAI}.

\bibitem[{Budzianowski et~al.(2018)Budzianowski, Wen, Tseng, Casanueva, Ultes,
  Ramadan, and Gasic}]{budzianowski2018multiwoz}
Pawe{\l} Budzianowski, Tsung-Hsien Wen, Bo-Hsiang Tseng, I{\~n}igo Casanueva,
  Stefan Ultes, Osman Ramadan, and Milica Gasic. 2018.
\newblock Multiwoz-a large-scale multi-domain wizard-of-oz dataset for
  task-oriented dialogue modelling.
\newblock In \emph{Proceedings of the 2018 Conference on Empirical Methods in
  Natural Language Processing}, pages 5016--5026.

\bibitem[{Chen et~al.(2017)Chen, Liu, Yin, and Tang}]{chen2017survey}
Hongshen Chen, Xiaorui Liu, Dawei Yin, and Jiliang Tang. 2017.
\newblock A survey on dialogue systems: Recent advances and new frontiers.
\newblock \emph{Acm Sigkdd Explorations Newsletter}, 19(2):25--35.

\bibitem[{Cui et~al.(2020)Cui, Wu, Liu, Zhang, and Zhou}]{cui2020mutual}
Leyang Cui, Yu~Wu, Shujie Liu, Yue Zhang, and Ming Zhou. 2020.
\newblock Mutual: A dataset for multi-turn dialogue reasoning.
\newblock In \emph{Proceedings of the 58th Annual Meeting of the Association
  for Computational Linguistics}, pages 1406--1416.

\bibitem[{Deriu et~al.(2020)Deriu, Rodrigo, Otegi, Echegoyen, Rosset, Agirre,
  and Cieliebak}]{deriu2020survey}
Jan Deriu, Alvaro Rodrigo, Arantxa Otegi, Guillermo Echegoyen, Sophie Rosset,
  Eneko Agirre, and Mark Cieliebak. 2020.
\newblock Survey on evaluation methods for dialogue systems.
\newblock \emph{Artificial Intelligence Review}, pages 1--56.

\bibitem[{Dinan et~al.(2018)Dinan, Roller, Shuster, Fan, Auli, and
  Weston}]{dinan2018wizard}
Emily Dinan, Stephen Roller, Kurt Shuster, Angela Fan, Michael Auli, and Jason
  Weston. 2018.
\newblock Wizard of wikipedia: Knowledge-powered conversational agents.
\newblock \emph{arXiv preprint arXiv:1811.01241}.

\bibitem[{Ghazvininejad et~al.(2018)Ghazvininejad, Brockett, Chang, Dolan, Gao,
  Yih, and Galley}]{ghazvininejad2018knowledge}
Marjan Ghazvininejad, Chris Brockett, Ming-Wei Chang, Bill Dolan, Jianfeng Gao,
  Wen-tau Yih, and Michel Galley. 2018.
\newblock A knowledge-grounded neural conversation model.
\newblock In \emph{Proceedings of the AAAI Conference on Artificial
  Intelligence}, volume~32.

\bibitem[{Gopalakrishnan et~al.(2019)Gopalakrishnan, Hedayatnia, Chen,
  Gottardi, Kwatra, Venkatesh, Gabriel, and
  Hakkani-T{\"u}r}]{gopalakrishnan2019topical}
Karthik Gopalakrishnan, Behnam Hedayatnia, Qinglang Chen, Anna Gottardi,
  Sanjeev Kwatra, Anu Venkatesh, Raefer Gabriel, and Dilek Hakkani-T{\"u}r.
  2019.
\newblock Topical-chat: Towards knowledge-grounded open-domain conversations.
\newblock In \emph{INTERSPEECH}, pages 1891--1895.

\bibitem[{Huang et~al.(2019)Huang, Bras, Bhagavatula, and
  Choi}]{huang2019cosmos}
Lifu Huang, Ronan~Le Bras, Chandra Bhagavatula, and Yejin Choi. 2019.
\newblock Cosmos qa: Machine reading comprehension with contextual commonsense
  reasoning.
\newblock \emph{arXiv preprint arXiv:1909.00277}.

\bibitem[{Li et~al.(2017)Li, Su, Shen, Li, Cao, and Niu}]{li2017dailydialog}
Yanran Li, Hui Su, Xiaoyu Shen, Wenjie Li, Ziqiang Cao, and Shuzi Niu. 2017.
\newblock Dailydialog: A manually labelled multi-turn dialogue dataset.
\newblock In \emph{Proceedings of the Eighth International Joint Conference on
  Natural Language Processing (Volume 1: Long Papers)}, pages 986--995.

\bibitem[{Lin et~al.(2020)Lin, Zhou, Shen, Zhou, Bhagavatula, Choi, and
  Ren}]{lin2020commongen}
Bill~Yuchen Lin, Wangchunshu Zhou, Ming Shen, Pei Zhou, Chandra Bhagavatula,
  Yejin Choi, and Xiang Ren. 2020.
\newblock Commongen: A constrained text generation challenge for generative
  commonsense reasoning.
\newblock In \emph{Proceedings of the 2020 Conference on Empirical Methods in
  Natural Language Processing: Findings}, pages 1823--1840.

\bibitem[{Lin(2004)}]{lin2004rouge}
Chin-Yew Lin. 2004.
\newblock Rouge: A package for automatic evaluation of summaries.
\newblock In \emph{Text summarization branches out}, pages 74--81.

\bibitem[{Liu et~al.(2016)Liu, Lowe, Serban, Noseworthy, Charlin, and
  Pineau}]{liu2016not}
Chia-Wei Liu, Ryan Lowe, Iulian~V Serban, Michael Noseworthy, Laurent Charlin,
  and Joelle Pineau. 2016.
\newblock How not to evaluate your dialogue system: An empirical study of
  unsupervised evaluation metrics for dialogue response generation.
\newblock \emph{arXiv preprint arXiv:1603.08023}.

\bibitem[{Liu and Singh(2004)}]{liu2004conceptnet}
Hugo Liu and Push Singh. 2004.
\newblock Conceptnet—a practical commonsense reasoning tool-kit.
\newblock \emph{BT technology journal}, 22(4):211--226.

\bibitem[{Lowe et~al.(2017)Lowe, Noseworthy, Serban, Angelard-Gontier, Bengio,
  and Pineau}]{lowe2017towards}
Ryan Lowe, Michael Noseworthy, Iulian~V Serban, Nicolas Angelard-Gontier,
  Yoshua Bengio, and Joelle Pineau. 2017.
\newblock Towards an automatic turing test: Learning to evaluate dialogue
  responses.
\newblock \emph{arXiv preprint arXiv:1708.07149}.

\bibitem[{Mehrabi et~al.(2021)Mehrabi, Zhou, Morstatter, Pujara, Ren, and
  Galstyan}]{mehrabi2021lawyers}
Ninareh Mehrabi, Pei Zhou, Fred Morstatter, Jay Pujara, Xiang Ren, and Aram
  Galstyan. 2021.
\newblock Lawyers are dishonest? quantifying representational harms in
  commonsense knowledge resources.
\newblock \emph{arXiv preprint arXiv:2103.11320}.

\bibitem[{Mehri and Eskenazi(2020{\natexlab{a}})}]{mehri2020unsupervised}
Shikib Mehri and Maxine Eskenazi. 2020{\natexlab{a}}.
\newblock Unsupervised evaluation of interactive dialog with dialogpt.
\newblock \emph{arXiv preprint arXiv:2006.12719}.

\bibitem[{Mehri and Eskenazi(2020{\natexlab{b}})}]{mehri2020usr}
Shikib Mehri and Maxine Eskenazi. 2020{\natexlab{b}}.
\newblock Usr: An unsupervised and reference free evaluation metric for dialog
  generation.
\newblock \emph{arXiv preprint arXiv:2005.00456}.

\bibitem[{Nie et~al.(2020)Nie, Williamson, Bansal, Kiela, and
  Weston}]{nie2020like}
Yixin Nie, Mary Williamson, Mohit Bansal, Douwe Kiela, and Jason Weston. 2020.
\newblock I like fish, especially dolphins: Addressing contradictions in
  dialogue modelling.
\newblock \emph{arXiv preprint arXiv:2012.13391}.

\bibitem[{Radford et~al.(2019)Radford, Wu, Child, Luan, Amodei, and
  Sutskever}]{radford2019language}
Alec Radford, Jeffrey Wu, Rewon Child, David Luan, Dario Amodei, and Ilya
  Sutskever. 2019.
\newblock Language models are unsupervised multitask learners.
\newblock \emph{OpenAI Blog 1.8 (2019): 9.}

\bibitem[{Rashkin et~al.(2019)Rashkin, Smith, Li, and
  Boureau}]{rashkin2019towards}
Hannah Rashkin, Eric~Michael Smith, Margaret Li, and Y-Lan Boureau. 2019.
\newblock Towards empathetic open-domain conversation models: A new benchmark
  and dataset.
\newblock In \emph{Proceedings of the 57th Annual Meeting of the Association
  for Computational Linguistics}, pages 5370--5381.

\bibitem[{Richardson et~al.(2020)Richardson, Hu, Moss, and
  Sabharwal}]{richardson2020probing}
Kyle Richardson, Hai Hu, Lawrence~S Moss, and Ashish Sabharwal. 2020.
\newblock Probing natural language inference models through semantic fragments.
\newblock In \emph{AAAI}, pages 8713--8721.

\bibitem[{Richardson and Sabharwal(2020)}]{richardson2020does}
Kyle Richardson and Ashish Sabharwal. 2020.
\newblock What does my qa model know? devising controlled probes using expert
  knowledge.
\newblock \emph{Transactions of the Association for Computational Linguistics},
  8:572--588.

\bibitem[{Roller et~al.(2020)Roller, Dinan, Goyal, Ju, Williamson, Liu, Xu,
  Ott, Shuster, Smith et~al.}]{roller2020recipes}
Stephen Roller, Emily Dinan, Naman Goyal, Da~Ju, Mary Williamson, Yinhan Liu,
  Jing Xu, Myle Ott, Kurt Shuster, Eric~M Smith, et~al. 2020.
\newblock Recipes for building an open-domain chatbot.
\newblock \emph{arXiv preprint arXiv:2004.13637}.

\bibitem[{Sap et~al.(2019{\natexlab{a}})Sap, Le~Bras, Allaway, Bhagavatula,
  Lourie, Rashkin, Roof, Smith, and Choi}]{sap2019atomic}
Maarten Sap, Ronan Le~Bras, Emily Allaway, Chandra Bhagavatula, Nicholas
  Lourie, Hannah Rashkin, Brendan Roof, Noah~A Smith, and Yejin Choi.
  2019{\natexlab{a}}.
\newblock Atomic: An atlas of machine commonsense for if-then reasoning.
\newblock In \emph{Proceedings of the AAAI Conference on Artificial
  Intelligence}, volume~33, pages 3027--3035.

\bibitem[{Sap et~al.(2019{\natexlab{b}})Sap, Rashkin, Chen, Le~Bras, and
  Choi}]{sap2019social}
Maarten Sap, Hannah Rashkin, Derek Chen, Ronan Le~Bras, and Yejin Choi.
  2019{\natexlab{b}}.
\newblock Social iqa: Commonsense reasoning about social interactions.
\newblock In \emph{Proceedings of the 2019 Conference on Empirical Methods in
  Natural Language Processing and the 9th International Joint Conference on
  Natural Language Processing (EMNLP-IJCNLP)}, pages 4453--4463.

\bibitem[{Sap et~al.(2020)Sap, Shwartz, Bosselut, Choi, and
  Roth}]{sap2020commonsense}
Maarten Sap, Vered Shwartz, Antoine Bosselut, Yejin Choi, and Dan Roth. 2020.
\newblock Commonsense reasoning for natural language processing.
\newblock In \emph{Proceedings of the 58th Annual Meeting of the Association
  for Computational Linguistics: Tutorial Abstracts}, pages 27--33.

\bibitem[{Sellam et~al.(2020)Sellam, Das, and Parikh}]{sellam2020bleurt}
Thibault Sellam, Dipanjan Das, and Ankur~P Parikh. 2020.
\newblock Bleurt: Learning robust metrics for text generation.
\newblock \emph{arXiv preprint arXiv:2004.04696}.

\bibitem[{Serban et~al.(2016)Serban, Sordoni, Bengio, Courville, and
  Pineau}]{serban2016building}
Iulian~V Serban, Alessandro Sordoni, Yoshua Bengio, Aaron Courville, and Joelle
  Pineau. 2016.
\newblock Building end-to-end dialogue systems using generative hierarchical
  neural network models.
\newblock In \emph{Thirtieth AAAI Conference on Artificial Intelligence}.

\bibitem[{Serban et~al.(2017)Serban, Sordoni, Lowe, Charlin, Pineau, Courville,
  and Bengio}]{serban2017hierarchical}
Iulian~Vlad Serban, Alessandro Sordoni, Ryan Lowe, Laurent Charlin, Joelle
  Pineau, Aaron Courville, and Yoshua Bengio. 2017.
\newblock A hierarchical latent variable encoder-decoder model for generating
  dialogues.
\newblock In \emph{Thirty-First AAAI Conference on Artificial Intelligence}.

\bibitem[{Sordoni et~al.(2015)Sordoni, Galley, Auli, Brockett, Ji, Mitchell,
  Nie, Gao, and Dolan}]{sordoni2015neural}
Alessandro Sordoni, Michel Galley, Michael Auli, Chris Brockett, Yangfeng Ji,
  Margaret Mitchell, Jian-Yun Nie, Jianfeng Gao, and Bill Dolan. 2015.
\newblock A neural network approach to context-sensitive generation of
  conversational responses.
\newblock \emph{arXiv preprint arXiv:1506.06714}.

\bibitem[{Talmor et~al.(2019)Talmor, Herzig, Lourie, and
  Berant}]{talmor2019commonsenseqa}
Alon Talmor, Jonathan Herzig, Nicholas Lourie, and Jonathan Berant. 2019.
\newblock Commonsenseqa: A question answering challenge targeting commonsense
  knowledge.
\newblock In \emph{Proceedings of the 2019 Conference of the North American
  Chapter of the Association for Computational Linguistics: Human Language
  Technologies, Volume 1 (Long and Short Papers)}, pages 4149--4158.

\bibitem[{Tao et~al.(2018)Tao, Mou, Zhao, and Yan}]{tao2018ruber}
Chongyang Tao, Lili Mou, Dongyan Zhao, and Rui Yan. 2018.
\newblock Ruber: An unsupervised method for automatic evaluation of open-domain
  dialog systems.
\newblock In \emph{Proceedings of the AAAI Conference on Artificial
  Intelligence}, volume~32.

\bibitem[{Vinyals and Le(2015)}]{vinyals2015neural}
Oriol Vinyals and Quoc Le. 2015.
\newblock A neural conversational model.
\newblock \emph{arXiv preprint arXiv:1506.05869}.

\bibitem[{Williams et~al.(2018)Williams, Nangia, and Bowman}]{MNLI}
Adina Williams, Nikita Nangia, and Samuel Bowman. 2018.
\newblock A broad-coverage challenge corpus for sentence understanding through
  inference.
\newblock In \emph{Proceedings of the 2018 Conference of the North American
  Chapter of the Association for Computational Linguistics: Human Language
  Technologies, Volume 1 (Long Papers)}, pages 1112--1122. Association for
  Computational Linguistics.

\bibitem[{Wolf et~al.(2019)Wolf, Sanh, Chaumond, and
  Delangue}]{transfertransfo}
Thomas Wolf, Victor Sanh, Julien Chaumond, and Clement Delangue. 2019.
\newblock Transfertransfo: {A} transfer learning approach for neural network
  based conversational agents.
\newblock \emph{CoRR}.

\bibitem[{Yu et~al.(2020)Yu, Jiang, Dong, and Feng}]{yu2020reclor}
Weihao Yu, Zihang Jiang, Yanfei Dong, and Jiashi Feng. 2020.
\newblock Reclor: A reading comprehension dataset requiring logical reasoning.
\newblock \emph{arXiv preprint arXiv:2002.04326}.

\bibitem[{Zellers et~al.(2018)Zellers, Bisk, Schwartz, and
  Choi}]{zellers2018swagaf}
Rowan Zellers, Yonatan Bisk, Roy Schwartz, and Yejin Choi. 2018.
\newblock Swag: A large-scale adversarial dataset for grounded commonsense
  inference.
\newblock In \emph{Proceedings of the 2018 Conference on Empirical Methods in
  Natural Language Processing (EMNLP)}.

\bibitem[{Zhang et~al.(2018)Zhang, Dinan, Urbanek, Szlam, Kiela, and
  Weston}]{zhang2018personalizing}
Saizheng Zhang, Emily Dinan, Jack Urbanek, Arthur Szlam, Douwe Kiela, and Jason
  Weston. 2018.
\newblock Personalizing dialogue agents: I have a dog, do you have pets too?
\newblock In \emph{Proceedings of the 56th Annual Meeting of the Association
  for Computational Linguistics (Volume 1: Long Papers)}, pages 2204--2213.

\bibitem[{Zhang et~al.(2019)Zhang, Kishore, Wu, Weinberger, and
  Artzi}]{zhang2019bertscore}
Tianyi Zhang, Varsha Kishore, Felix Wu, Kilian~Q Weinberger, and Yoav Artzi.
  2019.
\newblock Bertscore: Evaluating text generation with bert.
\newblock \emph{arXiv preprint arXiv:1904.09675}.

\bibitem[{Zhang et~al.(2020)Zhang, Sun, Galley, Chen, Brockett, Gao, Gao, Liu,
  and Dolan}]{zhang2020dialogpt}
Yizhe Zhang, Siqi Sun, Michel Galley, Yen-Chun Chen, Chris Brockett, Xiang Gao,
  Jianfeng Gao, Jingjing Liu, and William~B Dolan. 2020.
\newblock Dialogpt: Large-scale generative pre-training for conversational
  response generation.
\newblock In \emph{Proceedings of the 58th Annual Meeting of the Association
  for Computational Linguistics: System Demonstrations}, pages 270--278.

\bibitem[{Zhou et~al.(2018)Zhou, Young, Huang, Zhao, Xu, and
  Zhu}]{zhou2018commonsense}
Hao Zhou, Tom Young, Minlie Huang, Haizhou Zhao, Jingfang Xu, and Xiaoyan Zhu.
  2018.
\newblock Commonsense knowledge aware conversation generation with graph
  attention.
\newblock In \emph{IJCAI}, pages 4623--4629.

\bibitem[{Zhou et~al.(2020)Zhou, Khanna, Lee, Lin, Ho, Pujara, and
  Ren}]{zhou2020rica}
Pei Zhou, Rahul Khanna, Seyeon Lee, Bill~Yuchen Lin, Daniel Ho, Jay Pujara, and
  Xiang Ren. 2020.
\newblock Rica: Evaluating robust inference capabilities based on commonsense
  axioms.
\newblock \emph{arXiv preprint arXiv:2005.00782}.

\end{thebibliography}
\bibliographystyle{acl_natbib}

\appendix
\section{Example Responses and Scores}
\label{sec:appendix}
Table 5 shows responses to some dialogue histories from both human and models. We also include scores from human annotations described in \ref{sec:human eval}.
\begin{table*}[h]
\resizebox{\textwidth}{!}{
\begin{tabular}{|c|c|l|l|l|}
\hline
\multicolumn{2}{|c|}{History}                                                                                                                         & \begin{tabular}[c]{@{}l@{}}A: my pet dog \\ is always there for me. \\ No matter what, \\ I can count on her.\end{tabular}                                                              & \begin{tabular}[c]{@{}l@{}}A: james , why are \\ you watching tv ? \\ your train leaves \\ at 4:30. there are just \\ 2 hours left . \\ have you finished packing ?\end{tabular} & \begin{tabular}[c]{@{}l@{}}A: i am up for\\  a promotion at my job. \\ I positive I will get it. \\ B: That is great! \\ you should celebrate\\  and enjoy this moment\end{tabular} \\ \hline
\multirow{7}{*}{\begin{tabular}[c]{@{}c@{}}Human and Model  \\ Responses with \\ Annotation Scores \\ (1-10)\end{tabular}} & Human Response           & \begin{tabular}[c]{@{}l@{}}B: I feel the same \\ way about my dog. \\ They're the greatest! {[}10{]}\end{tabular}                                                                       & \begin{tabular}[c]{@{}l@{}}B: i 've almost \\ finished packing . now \\ i 'm waiting for the \\ jacket i lent to david \\ last week . {[}9{]}\end{tabular}                       & \begin{tabular}[c]{@{}l@{}}A: I already have plans\\  to go out to a \\ nice restaurant and \\ celebrate with my family {[}10{]}\end{tabular}                                       \\ \cline{2-5} 
                                                                                                                           & DD                       & B: I hope so. {[}3{]}                                                                                                                                                                   & \begin{tabular}[c]{@{}l@{}}B: yes, I've already \\ got the files ready. {[}1{]}\end{tabular}                                                                                     & A: You are so wonderful! {[}3{]}                                                                                                                                                    \\ \cline{2-5} 
                                                                                                                           & ED                       & \begin{tabular}[c]{@{}l@{}}B: That is so sweet. \\ How long has she \\ been with you? {[}10{]}\end{tabular}                                                                             & \begin{tabular}[c]{@{}l@{}}B: i hope you get \\ through it {[}1{]}\end{tabular}                                                                                                  & \begin{tabular}[c]{@{}l@{}}A: yeah i know, it is \\ such a blessing to\\  have a mom. {[}1{]}\end{tabular}                                                                          \\ \cline{2-5} 
                                                                                                                           & TC                       & \begin{tabular}[c]{@{}l@{}}B: I have a dog, \\ and it is a good \\ thing that I don't \\ have a cat, because \\ I am afraid of them. \\ My dog is a border collie. {[}2{]}\end{tabular} & \begin{tabular}[c]{@{}l@{}}B: I am watching tv, \\ just to catch up on \\ some shows, how \\ about you? {[}2{]}\end{tabular}                                                     & \begin{tabular}[c]{@{}l@{}}A: I love when my kids\\  get something too!\\  My kids have been \\ very successful and \\ I love how they \\ have friends. {[}2{]}\end{tabular}        \\ \cline{2-5} 
                                                                                                                           & Filtered Existing (FE)   & \begin{tabular}[c]{@{}l@{}}B: That's great. \\ You must miss her. {[}10{]}\end{tabular}                                                                                                 & \begin{tabular}[c]{@{}l@{}}B: i'm not sure. \\ i'm going to the movies. \\ i'm not sure if i 'll have\\  time to finish packing. {[}3{]}\end{tabular}                            & \begin{tabular}[c]{@{}l@{}}A: i am so happy for\\  my promotion. {[}4{]}\end{tabular}                                                                                               \\ \cline{2-5} 
                                                                                                                           & FE+Crowdsourced          & \begin{tabular}[c]{@{}l@{}}B: That is nice. \\ I have two dogs. {[}7{]}\end{tabular}                                                                                                    & \begin{tabular}[c]{@{}l@{}}B: oh, yes. i have. \\ i 'll be home at 7:30. {[}6{]}\end{tabular}                                                                                    & \begin{tabular}[c]{@{}l@{}}A: I hope so. I am so \\ excited about it. {[}7{]}\end{tabular}                                                                                          \\ \cline{2-5} 
                                                                                                                           & FE+Filtered Crowdsourced & \begin{tabular}[c]{@{}l@{}}B: That's nice. \\ What breed is it? {[}10{]}\end{tabular}                                                                                                   & \begin{tabular}[c]{@{}l@{}}B: oh, i finished \\ packing. i 'll go to \\ the station and \\ get my things. {[}9{]}\end{tabular}                                                   & \begin{tabular}[c]{@{}l@{}}A: Thanks! I am going\\ to get a new car! {[}8{]}\end{tabular}                                                                                           \\ \hline
\end{tabular}
}
\caption{Example responses of human and GPT2 trained on different data with scores from human annotations in brackets, ranging from 1 to 10.}
\label{tab:example_responses}
\end{table*}

\end{document}